\documentclass{ecai} 

\usepackage{latexsym}
\usepackage{amssymb}
\usepackage{amsmath}
\usepackage{amsthm}
\usepackage{booktabs}
\usepackage{enumitem}
\usepackage{graphicx}
\usepackage{color}
\usepackage{subcaption}
\usepackage{threeparttable}

\newcommand{\BibTeX}{B\kern-.05em{\sc i\kern-.025em b}\kern-.08em\TeX}
\newcommand{\fst}[1]{\textcolor{red}{\textbf{#1}}}
\newcommand{\snd}[1]{\textcolor{blue}{\textbf{#1}}}

\begin{document}


\begin{frontmatter}


\paperid{1311} 


\title{SFDFusion: An Efficient Spatial-Frequency Domain Fusion Network for Infrared and Visible Image Fusion}



\author[A]{\fnms{Kun}~\snm{Hu}\thanks{Corresponding Author. Email:kunhu@buaa.edu.cn.}}
\author[A]{\fnms{Qingle}~\snm{Zhang}}
\author[B]{\fnms{Maoxun}~\snm{Yuan}}
\author[A]{\fnms{Yitian}~\snm{Zhang}}
\address[A]{Institute of Artificial Intelligence, Beihang University, China} 
\address[B]{School of Computer Science and Engineering, Beihang University, China}

\begin{abstract}
Infrared and visible image fusion aims to utilize the complementary information from two modalities to generate fused images with prominent targets and rich texture details. Most existing algorithms only perform pixel-level or feature-level fusion from different modalities in the spatial domain. They usually overlook the information in the frequency domain, and some of them suffer from inefficiency due to excessively complex structures. To tackle these challenges, this paper proposes an efficient Spatial-Frequency Domain Fusion (SFDFusion) network for infrared and visible image fusion. First, we propose a Dual-Modality Refinement Module (DMRM) to extract complementary information. This module extracts useful information from both the infrared and visible modalities in the spatial domain and enhances fine-grained spatial details. Next, to introduce frequency domain information, we construct a Frequency Domain Fusion Module (FDFM) that transforms the spatial domain to the frequency domain through Fast Fourier Transform (FFT) and then integrates frequency domain information. Additionally, we design a frequency domain fusion loss to provide guidance for the fusion process. Extensive experiments on public datasets demonstrate that our method produces fused images with significant advantages in various fusion metrics and visual effects. Furthermore, our method demonstrates high efficiency in image fusion and good performance on downstream detection tasks, thereby satisfying the real-time demands of advanced visual tasks. The code is available at \url{https://github.com/lqz2/SFDFusion}.
\end{abstract}

\end{frontmatter}


\section{Introduction}

Image fusion aims to generate a more informative image from images captured by multi-source optical sensors \citep{li2017pixel,zhang2021deep,karim2023current}. Currently, image fusion is widely applied in various fields \citep{ma2019infrared}, such as remote sensing, autonomous driving, medical imaging diagnosis, and so on. Infrared and visible image fusion is a common task within the field of image fusion. Infrared images contain thermal radiation information. Even under conditions of limited illumination, the targets still have higher contrast in infrared images, making it easier to be identified \citep{li2013image}. However, the main disadvantage of infrared images is lack of texture details. Visible images, on the other hand, possess rich texture details but are greatly influenced by weather and illumination conditions \citep{li2019illumination}, causing loss of structural information in low-visibility scenarios. This disadvantage is highly detrimental to some downstream visual tasks. Infrared and visible image fusion leverages the imaging characteristics of both modalities to explore complementary information \citep{zhang2021image}. Thus, it can generate fused images with rich texture and contrast information, thereby enhancing the ability of human and machine visual perception.

\begin{figure}[t]
\centering
\includegraphics[width=0.99\linewidth]{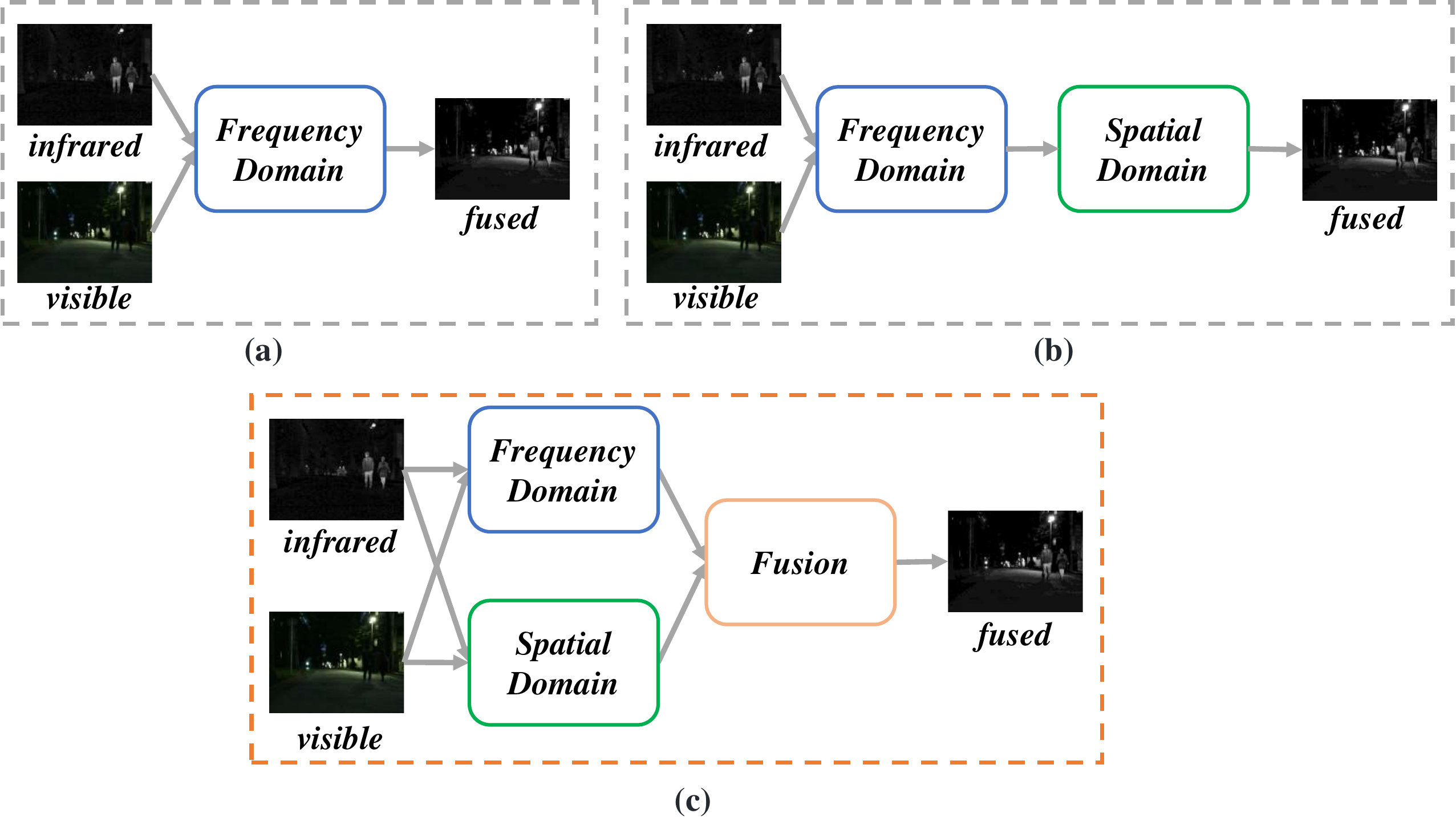}

\vspace{0.3cm}

\caption{Comparison of different methods for combining frequency domains. (a) Only frequency domain processing. (b) Serial structure for spatial and frequency domain processing. (c) Our parallel fusion structure of spatial and frequency domains.}

\vspace{0.55cm}

\label{fig1}
\end{figure}

The primary challenge in infrared and visible image fusion is how to extract valuable information from both modalities and utilize this information to generate a fused image. Early traditional methods \citep{li2020fast,zhou2016perceptual} mainly include sparse representation methods \citep{liu2016image}, multi-scale transform methods \citep{chen2020infrared}, subspace methods \citep{cvejic2007region}, and hybrid models \citep{ma2017infrared}. Although these traditional methods can generate decent fused images in some cases, the complexity of fusion rules \citep{yan2022injected} and the limitations of manual feature extraction restrict the expression of infrared and visible features, thereby reducing fusion performance \citep{liu2018deep}. With the success of deep learning-based methods in numerous computer vision tasks, some infrared and visible image fusion algorithms utilize Convolutional Neural Networks (CNNs) \citep{zhang2020ifcnn}, Generative Adversarial Networks (GANs) \citep{ma2019fusiongan,rao2023gan}, or Transformers \citep{ma2022swinfusion,tang2023datfuse} to obtain fused images. While these deep learning-based algorithms can produce satisfactory fused images, they mainly focus on spatial domain fusion \citep{zhang2021image}, neglecting information in the frequency domain. Moreover, some methods have highly complex network structures, resulting in slow inference speeds and difficulty in meeting the real-time requirements of practical scenarios \citep{xu2021drf}.

The image fusion methods mentioned above only focus on the image features in spatial domain, neglecting the features in frequency domain. In the spatial domain, the target and background can be better observed. In the frequency domain, edges and textures can be better identified by searching high-frequency information and repetitive patterns. Therefore, if a model fuses information from different domains, it will have a more comprehensive ability to understand images. Wang et al. \citep{wang2024efficient} decompose the input image into average coefficients and a set of high-frequency coefficients using Two-Dimensional Discrete Wavelet Transform (2D-DWT). They then build two fusion models to respectively fuse the average and high-frequency coefficients. Unfortunately, their network only captures frequency domain information during the learning process without incorporating spatial information, as illustrated in Figure~\ref{fig1} (a). Zheng et al. \citep{zheng2024frequency} conduct rigorous experimental analyses to demonstrate the feasibility of infrared and visible image fusion in the frequency domain. They utilize a frequency domain integration module to obtain the phase information of source images in the frequency domain for extracting salient targets. Furthermore, they improve texture details by reconstructing spatial domain images from the frequency domain using a spatial compensation module. The final fusion image is generated by a sequential processing of frequency and spatial domains, as illustrated in Figure~\ref{fig1} (b). However, a limitation of this serial structure is that the network may easily overlook valuable information from one domain while focusing on the effective information of the other domain.

Our motivation is that solely focusing on either the spatial domain or the frequency domain can result in the model failing to capture valuable information from the other domain. Additionally, processing these domains sequentially makes it challenging for the model to integrate complementary information from both domains. Therefore, the model should have the capability to learn hidden representations simultaneously from both the frequency and spatial domains. Hence, we design a parallel structure to fully extract valuable information from different domains, as shown in Figure~\ref{fig1} (c). 

Specifically, our paper introduces an efficient Spatial-Frequency Domain Fusion (SFDFusion) network tailored for fusing infrared and visible images. Our approach adopts a dual-branch strategy to capture valuable information from different modalities. We utilize a Dual-Modality Refinement Module (DMRM) to extract complementary information from both modalities in the spatial domain, enhancing gradient information. Moreover, we design a Frequency Domain Fusion Module (FDFM) where we convert the spatial domain to the frequency domain using Fast Fourier Transform (FFT) to fuse frequency domain information, then revert it back to the spatial domain using Inverse Fast Fourier Transform (IFFT). To ensure consistency between the frequency domain fusion results and the spatial domain images, we introduce a frequency domain fusion loss to guide the fusion process. Finally, the outputs from FDFM and DMRM are input into a simple fusion module to generate the final fused image. We compare our method with seven State-Of-The-Art (SOTA) methods on three public datasets. The experimental results indicate that the fused images generated by our method achieve better performance in various fusion metrics and visual effects.

In summary, the main contributions of this paper are as follows:
\begin{itemize}
    \item We propose an efficient infrared and visible image fusion network named SFDFusion. It can capture valuable information in the spatial domain, and supplement spatial domain information through the fusion of frequency domain information.
    \item We design the DMRM to eliminate redundant information in the spatial domain while preserving gradient information and texture details. Additionally, we develop the FDFM to capture effective information from two modalities in the frequency domain. To ensure the consistency between the frequency domain fusion results and spatial domain images, we also design a frequency domain fusion loss to guide the fusion process in the frequency domain.
    \item Our method has been validated to generate high-quality fusion images through experiments on various datasets. The fused image effectively preserves the information from the source images, while maintaining a high level of consistency with human visual perception. Furthermore, our method is more efficient and can effectively improve the accuracy of object detection.
\end{itemize}

\section{Related Work}

\subsection{Traditional Image Fusion}
Traditional image fusion methods typically require manual feature extraction and the design of corresponding fusion rules to merge images from different modalities. These methods include multi-scale transform-based methods, saliency-based methods, sparse representation-based methods, subspace-based methods, optimization-based methods, etc. Multi-scale transform-based methods decompose the source image into multi-scale representations and then merge them according to specific fusion rules. Saliency-based methods primarily detect salient regions through saliency detection \citep{liu2017infrared,bavirisetti2016two}. They extract relevant information from infrared and visible images, and integrate them into the fusion image. Sparse representation-based methods often generate an overcomplete dictionary from a large number of high-quality images using joint sparse representation \citep{yu2011image} and convolutional sparse representation \citep{liu2016image}. They then obtain the sparse representation of the source image from the overcomplete dictionary and generate the fusion image based on fusion rules. Subspace-based methods capture the intrinsic structure of the source image by mapping high-dimensional inputs to low-dimensional subspaces, typically using methods such as Principal Component Analysis (PCA) \citep{li2016improved}, Independent Component Analysis (ICA) \citep{cvejic2007region}, and Non-negative Matrix Factorization (NMF) \citep{mou2013image} for dimensionality reduction before fusion. Optimization-based methods generally design an objective function based on the overall intensity and texture details of the image \citep{zhao2017fusion}, and then seek the desired result by minimizing this objective function. While traditional methods have shown good fusion results in some scenarios, the limitations of manual feature extraction and the increasingly complex transformation methods and fusion rules restrict the acquisition of effective information and fusion efficiency. Therefore, researchers are paying more attention to deep learning-based infrared and visible image fusion methods in recent years \citep{zhang2021image}.

\begin{figure*}[htbp]
\centering
\includegraphics[width=0.9\linewidth]{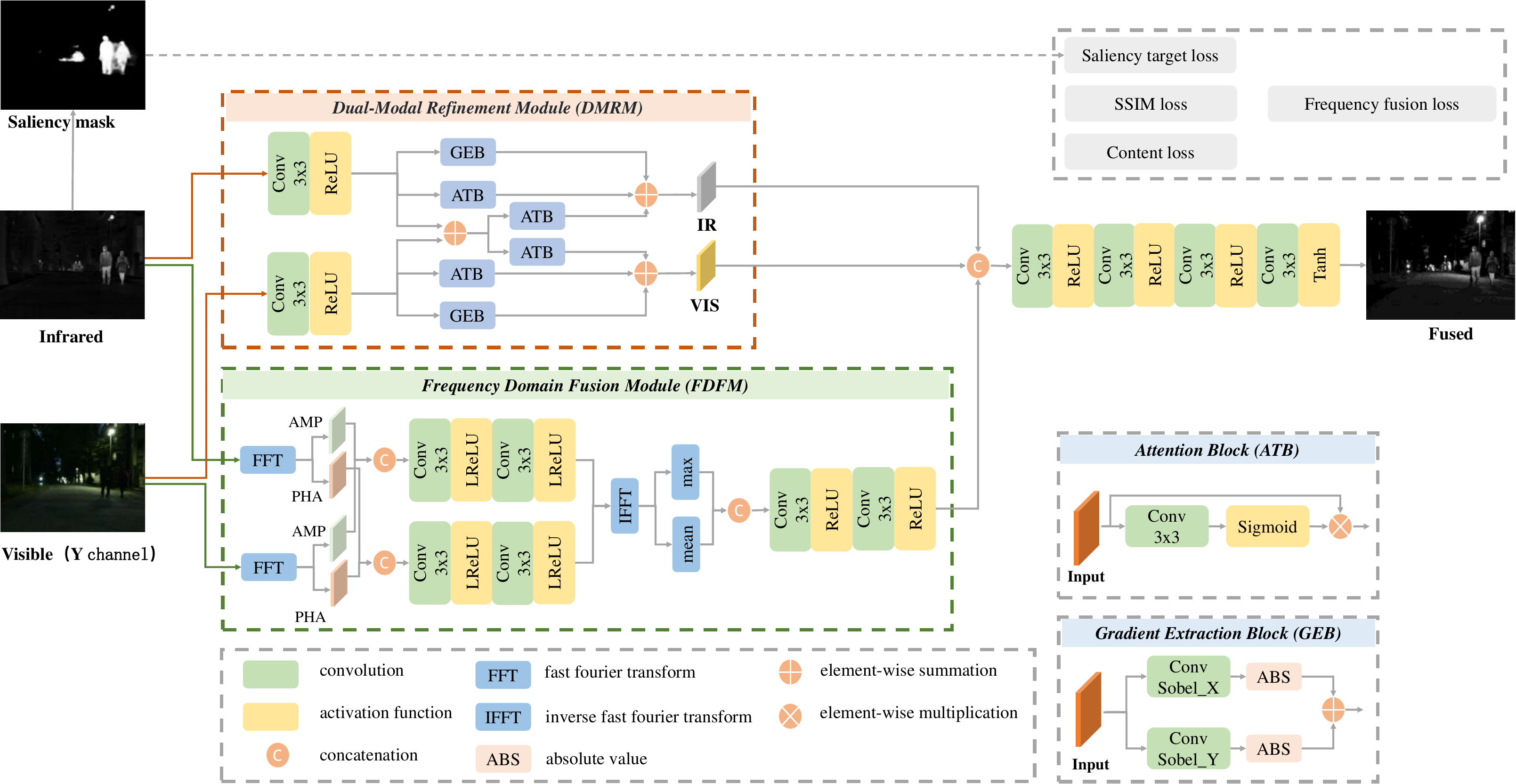}

\vspace{0.5cm}

\caption{The overall architecture of the SFDFusion network. The network adopts a parallel dual-branch structure, dedicated to refining the spatial domain and integrating the frequency domain. The spatial domain branch consists of DMRM, while the frequency domain branch consists of FDFM. After concatenating the outputs of different branches along the channel dimension, the final fused image is obtained through several simple convolution operations.}
\vspace{0.4cm}
\label{net}
\end{figure*}

\subsection{Deep Learning-Based Image Fusion}
Methods based on CNNs usually extract, integrate, or reconstruct implicit features of images. Researchers usually design refined loss functions to guide the network to generate high-quality fusion images containing complementary information. LP-CNN \citep{liu2017medical} is the first to apply CNN in the field of image fusion, using CNN for classification tasks and medical image fusion. Liu et al. \citep{liu2018infrared} obtain weight maps through CNN and completed fusion through image pyramids and multi-scale approaches. With the continuous development of CNNs, dense connections \citep{huang2017densely} are considered to be more effective in preserving shallow semantic information and enhancing information interaction within the network. Many image fusion methods begin to adopt dense structure. Li et al. \citep{li2018densefuse} incorporate the idea of dense connections by introducing dense blocks into the fusion network to reduce information loss during convolution processes. Xu et al. \citep{xu2020u2fusion} propose an unified fusion network U2Fusion based on dense networks and information metrics. It can adaptively estimate the importance of different modal images and incorporate the idea of continual learning to make the network adaptable to various fusion tasks. Furthermore, to fully retain significant target information, Ma et al. \citep{ma2021stdfusionnet} combine the traditional idea of saliency detection. Their method extracting saliency masks and incorporating them into the loss function to enhance the information of thermal radiation targets. Our method similarly adopts the idea of saliency detection. GANs can learn the distribution of real images to generate high-quality images, leading some image fusion algorithms to adopt GANs to obtain fusion images. Ma et al. \citep{ma2019fusiongan} first introduce GANs into the field of image fusion, supervising the generator to create texture-rich fusion images through the discriminator. Additionally, some GAN-based image fusion algorithms use dual discriminators to distinguish the differences between the fusion image and the two source images, such as DDcGAN \citep{ma2020ddcgan}. Transformers have achieved success in many visual tasks due to their self-attention mechanism that effectively addresses long-range dependencies in images. Wang et al. \citep{wang2022swinfuse} combine residual blocks with transformers to enhance fusion image quality by modeling long-range dependency relationships. Zhao et al. \citep{zhao2023cddfuse} construct a dual-branch transformer network to decompose and reconstruct features of source images and complete the image fusion task through the decoder. Different from above methods, our method simultaneously refines the spatial domain information and integrates the frequency domain information, overcoming the limitations of a single domain.


\section{Method}
\subsection{Overall Framework}
The overall framework of our network is shown in Figure 2. Given a pair of registered infrared image $I_{ir}\in\mathbb{R}^{1\times H\times W}$ and visible image $I_{vis}\in\mathbb{R}^{3\times H\times W}$, the visible image is first transformed from the RGB color space to the YCbCr color space, and the Y channel is extracted as the input of the visible image. We input infrared and visible images into DMRM to enhance and integrate key information and complementary information from different modalities, obtaining outputs $F_{ir}\in\mathbb{R}^{D\times H\times W}$ and $F_{vis}\in\mathbb{R}^{D\times H\times W}$, where D represents the dimension of embeddings. 

In the frequency domain branch, we first perform FFT on the infrared and visible images to obtain the amplitude and phase of the two modalities. Subsequently, the amplitude maps and phase maps of different modalities are concatenated along the channel dimension. The amplitude and phase are fused separately through convolutional blocks. Next, the fused amplitude and phase are transformed back to the spatial domain through IFFT. Afterwards, the maximum and average values are computed along the channel dimension from the previous output, then the results are concatenated and input into a convolutional block to obtain the output \(F_{fre}\in\mathbb{R}^{D\times H\times W}\) of FDFM.

Finally, concatenate $F_{ir}$, $F_{vis}$, and $F_{fre}$ along the channel dimension and feed them into a fusion block consisting of four $3 \times 3$ convolutional layers followed by activation functions. Through this process, we obtain the final fused image, with the formula as follow:
\begin{align}
    I_{fus}=Conv(\mathcal{C}(F_{ir},F_{vis},F_{fre}))
\end{align}
where $\mathcal{C}$ represents the concatenation operation in the channel dimension, and and $Conv$ denotes a convolutional block consisting of a convolutional layer and an activation function.



\subsection{Dual-Modality Refinement Module}
For infrared and visible images, a single modality often lacks crucial structural information of key targets or background texture information. To obtain higher quality fusion images, we refine the infrared and visible images through the DMRM module, which can also be viewed as a reconstruction process. Specifically, the DMRM module first expands the dimension to $\mathbb{R}^{D\times H\times W}$ using a convolutional layer and ReLU activation layer. The formula is as follow:
\begin{align}
\begin{Bmatrix}
    \hat{F}_{ir},\hat{F}_{vis}
\end{Bmatrix}=\begin{Bmatrix}\delta(Conv(I_{ir})),\delta(Conv(I_{vis}))\end{Bmatrix}
\end{align}
where $\delta(\cdot)$ represents the ReLU function.

For each modality, we design three branches to enhance the effective information, namely the texture information branch, attention branch, and complementary information branch. In the texture information branch, we extract gradient information from the images using Gradient Extraction Blocks (GEBs). GEBs convolve the images in the X and Y directions with Sobel operators, then take the absolute values of the gradient information in each direction and sum them. In the attention branch, we suppress redundant information expression through a simple Attention Block (ATB), consisting of a convolutional layer and a Sigmoid activation layer. We multiply the output of the activation layer by the input of the convolutional layer to obtain the output of the ATB. In the complementary information branch, we aim for the network to adaptively capture complementary information from other modalities. Therefore, we add the infrared and visible images element-wise, then learn the complementary information of different modalities through two ATB blocks separately. Finally, the outputs of the three branches for infrared and visible images are summed together to serve as the output of the DMRM module, as shown in the following formula:
\begin{align}
F_{ir} &=GEB(\hat{F}_{ir})\oplus ATB(\hat{F}_{ir})\oplus ATB(\hat{F}_{ir}\oplus \hat{F}_{vis}) \\
F_{vis} &=GEB(\hat{F}_{vis})\oplus ATB(\hat{F}_{vis})\oplus ATB(\hat{F}_{ir}\oplus \hat{F}_{vis})
\end{align}
where $\oplus$ refers to element-wise summation.

\subsection{Frequency Domain Fusion Module}
The frequency domain information of an image provides another perspective for exploring hidden image features. In the frequency domain, edges and textures can be better identified by searching for high-frequency information and repetitive patterns. Moreover, the global characteristic of the frequency domain partially compensates for the negative effects of the local characteristic of convolutions. Therefore, we propose a light-weighted module named FDFM to integrate frequency domain information.

In FDFM, we first transform the infrared and visible images from the spatial domain to the frequency domain through FFT, then we extract the amplitude and phase from the spectrum of the infrared and visible images respectively, as shown in the following formula:
\begin{align}
\begin{Bmatrix}
    AMP_{ir},AMP_{vis}
\end{Bmatrix} &=\begin{Bmatrix}|FFT(I_{ir})|,|FFT(I_{vis})|\end{Bmatrix} \\
\begin{Bmatrix}
    PHA_{ir},PHA_{vis}
\end{Bmatrix} &=\begin{Bmatrix}\theta(FFT(I_{ir})),\theta(FFT(I_{vis}))\end{Bmatrix}
\end{align}
where $|\cdot|$ represents the absolute value function, and $\theta$ denotes the element-wise angle.

Then, we construct two branches to individually fuse the amplitude and phase. Each branch, composed of a convolutional block consisting of two convolutional layers and two Leaky ReLU activation layers, captures and integrates effective information from different modalities. Subsequently, we perform IFFT on the fused result in the frequency domain, converting the output back from the frequency domain to the spatial domain. The entire process can be represented by the following formula:
\begin{equation}
    \begin{aligned}
    \hat{F}_{fre} =IFFT(&Conv(\mathcal{C}(AMP_{ir},AMP_{vis})), \\
    &Conv(\mathcal{C}(AMP_{ir},AMP_{vis})))
    \end{aligned}
\end{equation}

In order to enhance the performance of the frequency domain fusion branch and increase its output dimension, we first compute the maximum and mean values of the previous outputs along the channel dimension. These values are then concatenated along the channel dimension and passed through a simple convolutional block to derive the final output of FDFM, illustrated in the following formula:
\begin{align}
    F_{fre}=Conv(\mathcal{C}(max(\hat{F}_{fre}),mean(\hat{F}_{fre})))
\end{align}
where $max(\cdot)$ denotes the element-wise maximum selection, and $mean(\cdot)$ denotes the element-wise average operation.
\subsection{Loss Function}
The loss function determines the proportion of information retained from different modalities of images. It also aids in enhancing the visual quality and evaluation metrics of the fused images. In the experiments of this paper, we guide the network training process through a combination of content loss $\mathcal{L}_{content}$, structural similarity loss $\mathcal{L}_{SSIM}$, saliency target loss $\mathcal{L}_{saliency}$, and frequency domain fusion loss $\mathcal{L}_{fre}$. The total loss is represented as follow:
\begin{align}
   \label{total_loss} \mathcal{L}_{\mathrm{total}}=\mathcal{L}_{content}+\mathcal{L}_{SSIM}+\lambda_s\mathcal{L}_{saliency}+\mathcal{L}_{fre}
\end{align}
where $\lambda_s$ is the balancing coefficient.

To enhance the contrast information and texture details in the fused images, we utilize content losses, which primarily consist of pixel intensity loss and gradient loss. Their computation is as follow:
\begin{align}
    \mathcal{L}_{int} &=\frac{1}{HW}\|I_{fus}-max(I_{ir},I_{vis})\|_{1} \\
    \mathcal{L}_{grad} &=\frac1{HW}\||\nabla I_{fus}|-max(|\nabla I_{ir}|,|\nabla I_{vis}|)\|_1
\end{align}
where $H$ and $W$ are the height and width of an image, respectively, $\|\cdot\|_{1}$ stands for the $l_1\text{-norm}$,  $\nabla$ indicates the Sobel 
gradient operator. Therefore, the content loss can be represented as follow:
\begin{align}
    \label{loss_content}    \mathcal{L}_{content}=\alpha_1\mathcal{L}_{int}+\alpha_2\mathcal{L}_{grad}
\end{align}
where  $\alpha_1$ and $\alpha_2$ are the tuning parameters.

The Structural Similarity Index (SSIM) \citep{wang2004image} is one of the commonly used metrics for image quality assessment. It comprehensively evaluates the quality of an image in terms of brightness, contrast, and structure. Therefore, many image fusion algorithms take structural similarity index loss into account when designing loss functions. The calculation formula is as follow:
\begin{align}
    \mathcal{L}_{SSIM} = \frac{1 - SSIM(I_{fus},I_{ir})}{2} + \frac{1 - SSIM(I_{fus},I_{vis})}{2} 
\end{align}

To better extract salient target information from infrared images and texture information from visible images, we refer to the method in STDFusion \citep{ma2021stdfusionnet}. We utilize the pre-trained U2Net \citep{qin2020u2} to generate saliency masks from infrared images and calculate saliency target loss based on these masks. The calculation method is as follow:
\begin{equation}    
    \begin{aligned}
    \mathcal{L}_{saliency}=& \frac{\beta}{HW}\left\|{\mathcal M}\otimes I_{ir}-{\mathcal M}\otimes I_{fus}\right\|_{1}+  \\
    &\frac{1}{HW}\left\|(1-{\mathcal M})\otimes I_{vis}-(1-{\mathcal M})\otimes I_{fus}\right\|_{1}
    \end{aligned}
\end{equation}
where $\mathcal{M}$ represents the saliency mask, $\otimes$ denotes element-wise multiplication, and $\beta$ is the balancing coefficient.

During the fusion process in the frequency domain, we use simple convolutional layers to merge the amplitude and phase. Due to the global nature of the frequency domain, the fusion process may introduce additional noise or lose some pixel details due to convolution operations, potentially reducing the quality of the final fusion result. Therefore, we design a frequency domain fusion loss $\mathcal{L}_{fre}$ to guide the fusion process in the frequency domain. The loss aims to ensure the consistency between the frequency domain fusion results and the spatial domain source images, and to preserve complementary information from different modalities as much as possible. Specifically, we transform the fused amplitude and phase back to the spatial domain through IFFT, and then calculate their correlation coefficients (CC) with prominent thermal targets in the infrared image and background details in the visible image based on saliency masks. The calculation formula is as follow:
\begin{equation}
    \begin{aligned}
        \mathcal{L}_{fre}= & \mathcal{CC}(\mathcal{M}\otimes\mathcal{X},\mathcal{M}\otimes I_{ir})+ \\
        &\mathcal{CC}((1-\mathcal{M})\otimes\mathcal{X},(1-\mathcal{M})\otimes I_{vis})
    \end{aligned}   
\end{equation}

where $\mathcal{X}=\mathcal{F}^{-1}(I_{amp},I_{pha})$, $\mathcal{F}^{-1}$ denotes the Inverse Fast Fourier Transform (IFFT), $I_{amp}$ represents the fused amplitude image, and $I_{pha}$ represents the fused phase image. $\mathcal{M}$ represents the saliency mask, $\mathcal{CC}(\cdot,\cdot)$ is the correlation coefficient operator.

\section{Experiments}
\subsection{Datasets and Metrics}
We conduct experiments on three datasets: M3FD \citep{liu2022target}, MSRS \citep{Tang2022PIAFusion}, and RoadScene \citep{xu2020u2fusion}, which are commonly used in recent years for infrared and visible image fusion algorithms. 
\begin{itemize}
    \item \textbf{M3FD}: The dataset contains a total of 4200 aligned pairs of infrared and visible images with corresponding object detection labels. The images are of size $1024 \times 768$, with most scenes captured in university campuses and urban roads. Since the dataset is not pre-divided, we randomly select 840 pairs of infrared and visible images for testing, representing a split ratio of 0.2.
    \item \textbf{MSRS}: The dataset comprises 1444 aligned pairs of infrared and visible images, showcasing road scenes in both daytime and nighttime. These images, with a resolution of $640 \times 480$, are divided into 1083 pairs for training and 361 pairs for testing.
    \item \textbf{RoadScene}: This dataset is widely used in the task of infrared and visible image fusion, primarily consisting of 221 pairs of infrared and visible images selected from the FLIR dataset. These images are captured by vehicle-mounted cameras, already aligned but with varying resolutions.
\end{itemize}

We use six metrics \citep{ma2019infrared} to comprehensively evaluate the quality of image fusion, namely entropy (EN), standard deviation (SD), spatial frequency (SF), mutual information (MI), visual information fidelity (VIF), and Gradient-based fusion performance (Qabf). Higher values for these metrics indicate better fusion results. In addition, we utilize the mean Average Precision (mAP) as the evaluation metric for the object detection tasks.

\begin{table}[htbp]

    \centering
    \caption{Quantitative comparison results across three datasets.}
    
    \vspace{0.2cm}
    
    \label{tab1}
    \makebox[0.82\linewidth]{Results on M3FD dataset.} 
    
    \vspace{0.15cm}
    
    \resizebox{0.82\linewidth}{!}{
        \Large
        \begin{tabular}{ccccccc}
        \toprule
        Methods & EN & SD & SF & MI & VIF & Qabf\\
        \midrule
        DIDFuse & \fst{7.175} & \fst{46.526} & \fst{14.805} & \snd{3.970} & \snd{0.675} & 0.506\\
        U2Fusion & 6.853 & 33.092 & 12.942 & 3.665 & 0.641 & \snd{0.565}\\
        YDTR & 6.583 & 28.261 & 10.894 & 3.838 & 0.623 & 0.460\\
        MFEIF & 6.686 & 30.070 & 9.165 & 3.887 & 0.639 & 0.458\\
        UMF-CMGR & 6.697 & 31.009 & 9.242 & 3.891 & 0.592 & 0.396\\
        TarDal & \snd{7.169} & \snd{42.947} & 13.471 & 3.900 & 0.587 & 0.451\\
        IRFS & 6.738 & 31.505 & 11.270 & 3.719 & 0.618 & 0.515\\
        Ours & 6.840 & 35.639 & \snd{13.618} & \fst{4.686} & \fst{0.772} & \fst{0.571}\\
        \bottomrule
        \end{tabular}
    }

    \vspace{0.2cm}
    
    \makebox[0.82\linewidth]{Results on MSRS dataset.} 

    \vspace{0.15cm}

    \resizebox{0.82\linewidth}{!}{
        \Large
        \begin{tabular}{ccccccc}
        \toprule
        Methods & EN & SD & SF & MI & VIF & Qabf\\
        \midrule
        DIDFuse & 4.080 & 29.959 & 9.641 & 2.307 & 0.302 & 0.200\\
        U2Fusion & 5.297 & 24.222 & 8.608 & 2.439 & 0.529 & 0.451\\
        YDTR & 5.645 & 25.371 & 7.402 & 2.774 & 0.577 & 0.348\\
        MFEIF & 5.775 & 33.822 & 8.076 & \snd{3.006} & \snd{0.753} & \snd{0.548}\\
        UMF-CMGR & 5.598 & 20.755 & 7.105 & 2.451 & 0.453 & 0.265\\
        TarDal & \snd{6.476} & \snd{37.656} & \snd{10.752} & 2.878 & 0.694 & 0.452\\
        IRFS & 6.603 & 35.869 & 9.888 & 2.836 & 0.733 & 0.458\\
        Ours & \fst{6.670} & \fst{43.197} & \fst{11.070} & \fst{3.914} & \fst{1.011} & \fst{0.680}\\
        \bottomrule
        \end{tabular}
    }

    \vspace{0.2cm}
    
    \makebox[0.82\linewidth]{Results on RoadScene dataset.} 

    \vspace{0.15cm}
    
    \resizebox{0.82\linewidth}{!}{
        \Large
        \begin{threeparttable}
        \begin{tabular}{ccccccc}
        \toprule
        Methods & EN & SD & SF & MI & VIF & Qabf\\
        \midrule
        DIDFuse & \snd{7.380} & \snd{51.591} & \snd{14.134} & 3.868 & 0.579 & 0.457\\
        U2Fusion & 7.039 & 36.146 & 13.214 & 3.706 & 0.563 & \fst{0.521}\\
        YDTR & 6.914 & 35.428 & 10.569 & 3.754 & 0.553 & 0.452\\
        MFEIF & 7.047 & 38.708 & 9.243 & \snd{3.947} & 0.618 & 0.455\\
        UMF-CMGR & 7.074 & 37.515 & 10.283 & 3.929 & \snd{0.624} & \snd{0.459}\\
        TarDal & 7.354 & 48.060 & 13.653 & 4.140 & 0.539 & 0.452\\
        IRFS & 7.029 & 36.199 & 10.034 & 3.739 & 0.574 & 0.453\\
        Ours & \fst{7.445} & \fst{56.095} & \fst{15.305} & \fst{4.595} & \fst{0.781} & 0.457\\
        \bottomrule
        \end{tabular}

        \vspace{0.2cm}
        
        \begin{tablenotes}
            \item[*] Data labeled in \fst{red} indicates the best results, while \snd{blue} indicates the second best.
        \end{tablenotes}
        \end{threeparttable}
    }
\end{table}

\subsection{Implement Details}
In the experimental section of this paper, we conduct training on the MSRS dataset and testing on the M3FD, MSRS, and RoadScene datasets. The training is performed using an NVIDIA GeForce RTX 2080 Ti GPU. We set the batch size for images to 4 and trained for a total of 50 epochs. To efficiently optimize the network parameters, we utilize the Adam optimizer to optimize the network, where the parameters $\beta_1$ and $\beta_2$ are set to 0.9 and 0.999. The initial learning rate is set to 0.0005, and the hyperparameters $\lambda_s, \alpha_1, \alpha_2$, and $\beta$ in the loss function are set to 10, 5, 10, and 5, respectively.

\begin{figure*}[htbp]
    \centering
    \begin{subfigure}{0.18\linewidth}
        \centering
        \includegraphics[width=\linewidth]{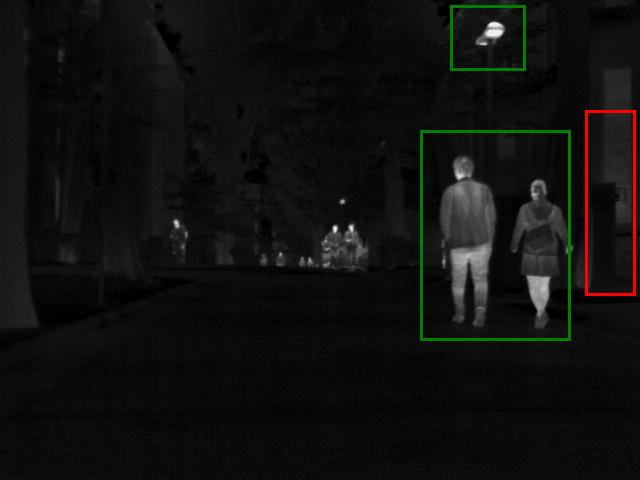}
        \caption{Infrared}
    \end{subfigure}
    \begin{subfigure}{0.18\linewidth}
        \centering
        \includegraphics[width=\linewidth]{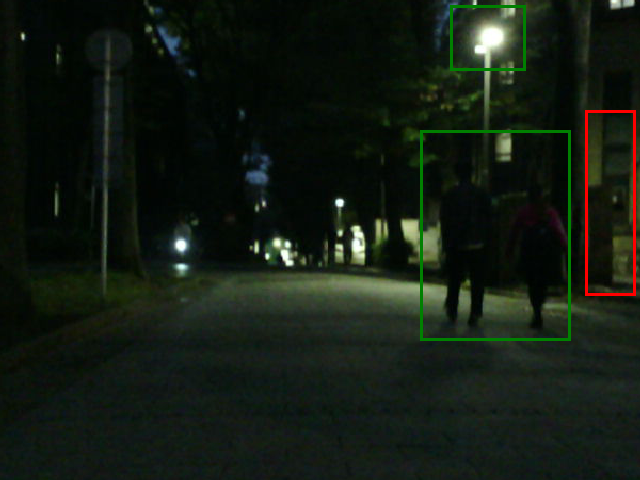}
        \caption{Visible}
    \end{subfigure}
    \begin{subfigure}{0.18\linewidth}
        \centering
        \includegraphics[width=\linewidth]{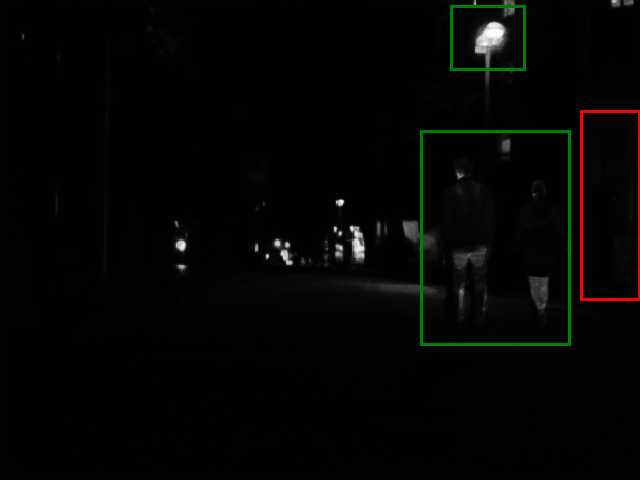}
        \caption{DIDFuse}
    \end{subfigure}
    \begin{subfigure}{0.18\linewidth}
        \centering
        \includegraphics[width=\linewidth]{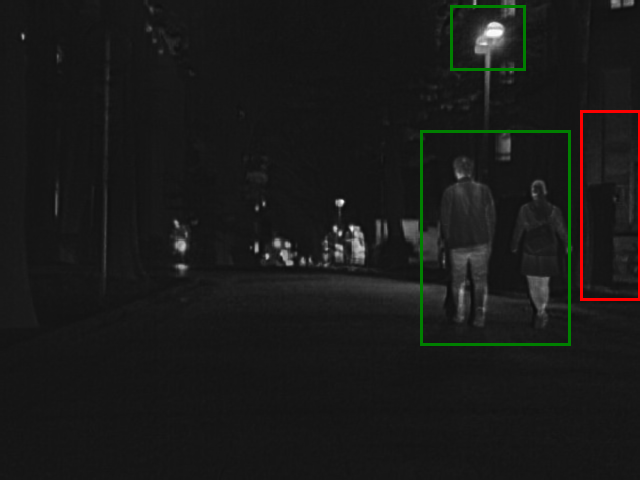}
        \caption{U2Fusion}
    \end{subfigure}
    \begin{subfigure}{0.18\linewidth}
        \centering
        \includegraphics[width=\linewidth]{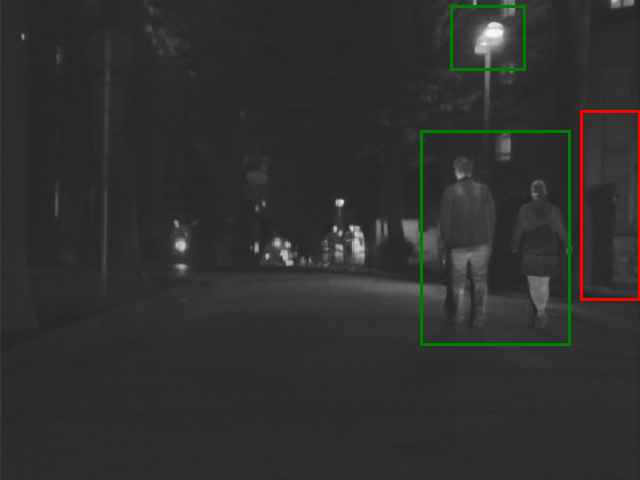}
        \caption{YDTR}
    \end{subfigure}

    \vspace{0.4cm}
    
    \begin{subfigure}{0.18\linewidth}
        \centering
        \includegraphics[width=\linewidth]{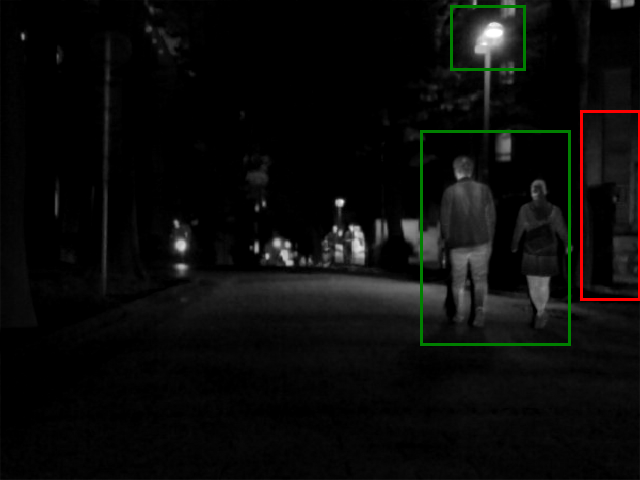}
        \caption{MFEIF}
    \end{subfigure}
    \begin{subfigure}{0.18\linewidth}
        \centering
        \includegraphics[width=\linewidth]{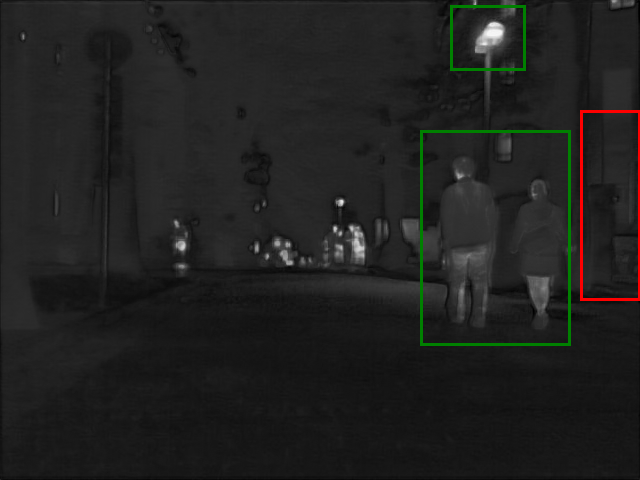}
        \caption{UMF-CMGR}
    \end{subfigure}
    \begin{subfigure}{0.18\linewidth}
        \centering
        \includegraphics[width=\linewidth]{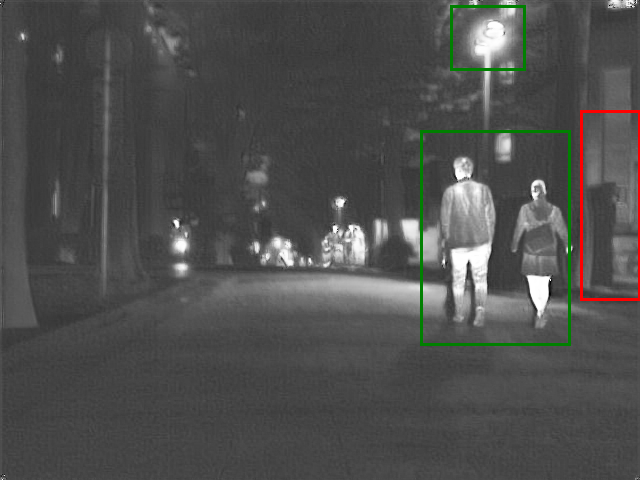}
        \caption{TarDal}
    \end{subfigure}
    \begin{subfigure}{0.18\linewidth}
        \centering
        \includegraphics[width=\linewidth]{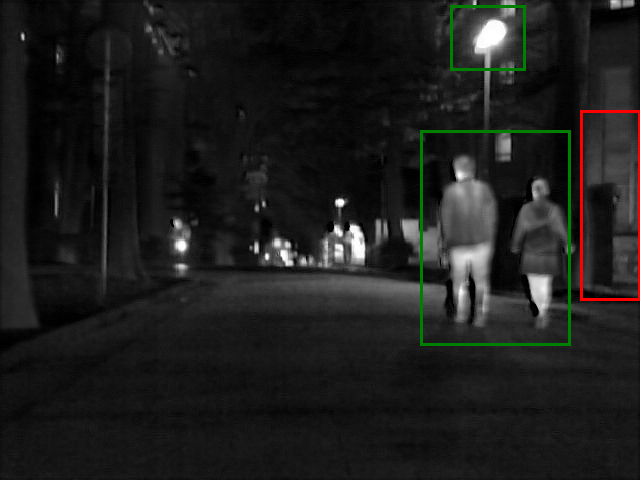}
        \caption{IRFS}
    \end{subfigure}
    \begin{subfigure}{0.18\linewidth}
        \centering
        \includegraphics[width=\linewidth]{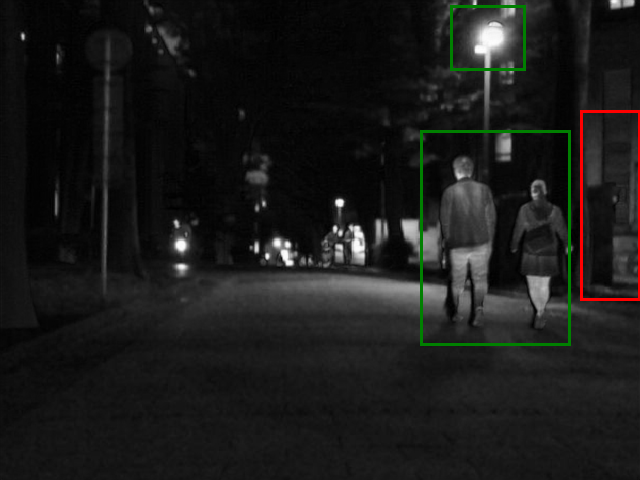}
        \caption{Ours}
    \end{subfigure}

    \vspace{0.4cm}
    
    \caption{Comparison of fusion results on the image "00024N" from the MSRS dataset using different methods.}

    \vspace{0.4cm}
    
    \label{fig4}
\end{figure*}

\subsection{Infrared and visible image fusion performance}
\subsubsection{Quantitative Analysis}
We compare our proposed method with seven SOTA image fusion methods to showcase its superiority. These methods include DIDFuse \citep{zhao2020didfuse}, U2Fusion \citep{xu2020u2fusion}, YDTR \citep{tang2022ydtr}, MFEIF \citep{liu2021learning}, UMF-CMGR \citep{Wang_2022_IJCAI}, TarDal \citep{liu2022target}, and IRFS \citep{wang2023interactively}. The results of this comparison can be found in Table~\ref{tab1}.

From the results in the table, it is evident that our method demonstrates excellent performance across all metrics on the MSRS dataset. While not achieving optimal results in all metrics on the M3FD and RoadScene datasets, our method excels in most metrics, showcasing a leading position. Overall, our method exhibits significant advantages in the MI and VIF metrics compared to other datasets. MI is typically used to indicate the degree of information preservation from the source images, a higher value indicates more preserved information from the source images. The superior performance in the MI metric suggests that our DMRM and FDFM modules effectively retain essential information from the source images, enhancing fusion quality. The VIF metric shows high consistency with human vision, where a larger VIF value indicates images that better meet human visual requirements. Our algorithm efficiently extracts complementary information from the infrared and visible modalities and incorporates saliency detection in the loss design. This integration results in fused images with more prominent foreground objects and richer texture details, aligning better with human visual perception habits.

In addition, we compare the inference time, parameter count, and computational load of SFDFusion with other methods, as shown in Table~\ref{tab-efficiency}. From the table, it is evident that our approach holds a significant advantage in terms of computational load. While we may not achieve the best performance in terms of inference time and parameter count among the compared methods, we still exhibit the second-best performance, with a very minor difference from the top-performing metric. The excellent efficiency indicates that our method holds great practical value in real-world scenarios.
\begin{table}[t]
\centering

\caption{Model efficiency comparison}
\label{tab-efficiency}
\renewcommand{\arraystretch}{0.9}
\setlength{\tabcolsep}{4mm}
\vspace{0.2cm}

\begin{tabular}{cccc}
\toprule
Method & Time (s) & Params/M & FLOPs/G \\
\midrule
DIDFuse & 0.043 & 0.26 & 114.86 \\
U2Fusion & 0.246 & 0.66 & 366.34 \\
YDTR & 0.033 & \fst{0.11} & 96.47 \\
MFEIF & 0.070 & 0.37 & 114.06 \\
UMFusion & 0.039 & 0.63 & 193.21 \\
TarDal & \fst{0.002} & 0.30 & 91.14 \\
IRFS & 0.002& 0.24 & \snd{73.74} \\
Ours & \snd{0.003} & \snd{0.14} & \fst{42.81} \\
\bottomrule
\end{tabular}

\end{table}
\subsubsection{Qualitative Analysis}
To visually compare the differences in fusion results obtained using different methods, we select a pair of typical infrared and visible images from the MSRS dataset and fused them using various methods, as shown in Figure~\ref{fig4}. We have highlighted the prominent foreground target regions with green boxes and the background texture regions with red boxes.

From the images, it is evident that our SFDFusion method effectively preserves the thermal radiation information from the infrared image for the foreground thermal targets. And the clear texture details of the targets are also well maintained. This is achieved by efficiently combining spatial and frequency domain information and enhancing the contrast of thermal targets through a saliency mask-related loss function. For the background textures, our method maximally retains most of the texture details from the visible image, primarily due to our DMRM module and gradient-related loss function. While TarDal and IRFS also emphasize the thermal targets, they still exists some problems. TarDal's result is affected by some degree of spectral contamination, leading to an overly bright background in the fusion image. On the other hand, although IFRS does not contain excessive redundant information in the background, the texture in the salient target region is excessively blurred. In comparison with other methods, our approach produces fusion images that strike a balance between highlighting foreground targets and preserving clear texture details, closely aligning with human visual perception.

\subsection{Object Detection Performance}
To evaluate the performance of image fusion methods on downstream tasks, we test the detection performance of different methods on the M3FD dataset. Specifically, we train YOLOv8 \citep{Jocher_Ultralytics_YOLO_2023} object detection networks on both infrared and visible images separately. Subsequently, we validate the fusion images generated by different methods using the detection networks trained on infrared modality and visible modality, respectively. For the detection results from different modalities, we consider the maximum value. The final results are shown in Table~\ref{tab2}.
\begin{table}[t]
\centering

\caption{Object detection metrics (mAP \text{@}0.5) on the M3FD dataset.}
\label{tab2}

\vspace{0.2cm}

\resizebox{0.99\linewidth}{!}{
    \begin{tabular}{cccccccc}
    \toprule
    \text{Methods} & \text{People} & \text{Car} & \text{Bus} & \text{Motorcycle} & \text{Lamp} & \text{Truck} & \text{All} \\
    \midrule
    DIDFuse & 0.460 & 0.867 & 0.631 & 0.607 & 0.530 & \snd{0.225} & 0.559 \\
    U2Fusion & 0.596 & 0.915 & \snd{0.716} & 0.762 & 0.548 & 0.995 & 0.755 \\
    YDTR & \snd{0.602} & 0.898 & 0.618 & 0.752 & 0.675 & 0.995 & 0.757 \\
    MFEIF & 0.582 & \snd{0.919} & 0.655 & 0.799 & \snd{0.675} & 0.995 & 0.771 \\
    UMFusion & 0.565 & 0.893 & \fst{0.718} & 0.733 & 0.572 & 0.995 & 0.746 \\
    TarDal & 0.556 & 0.878 & 0.630 & 0.701 & 0.547 & 0.995 & 0.718 \\
    IRFS & \fst{0.616} & 0.916 & 0.695 & \snd{0.815} & 0.642 & 0.995 & \snd{0.780} \\
    Ours & 0.590 & \fst{0.919} & 0.662 & \fst{0.875} & \fst{0.728} & \fst{0.995} & \fst{0.795} \\
    \bottomrule
    \end{tabular}
}
\end{table}

From the results, although our method is not specifically optimized for object detection tasks, it still maintains excellent detection performance on downstream object detection tasks while generating high-quality fusion images. This indicates that our network can learn more effective feature information during the fusion process and capture subtle features between different modalities.

\begin{figure*}[htbp]
    \centering
    \begin{subfigure}{0.18\linewidth}
        \centering
        \includegraphics[width=\linewidth]{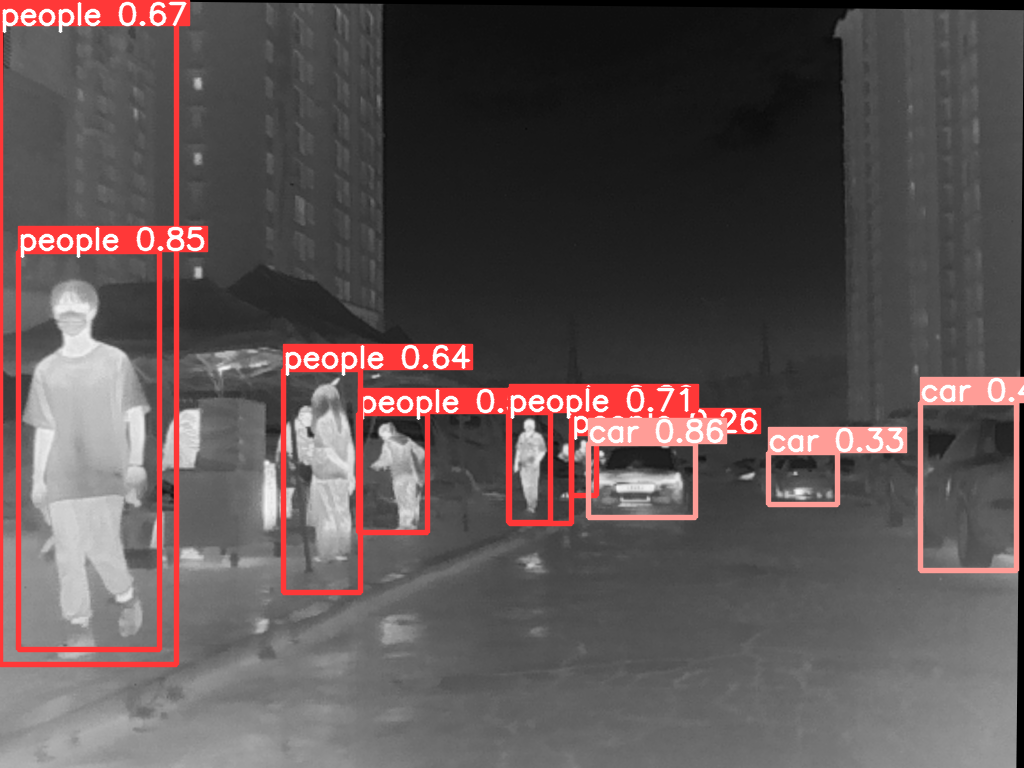}
        \caption{Infrared}
    \end{subfigure}
    \begin{subfigure}{0.18\linewidth}
        \centering
        \includegraphics[width=\linewidth]{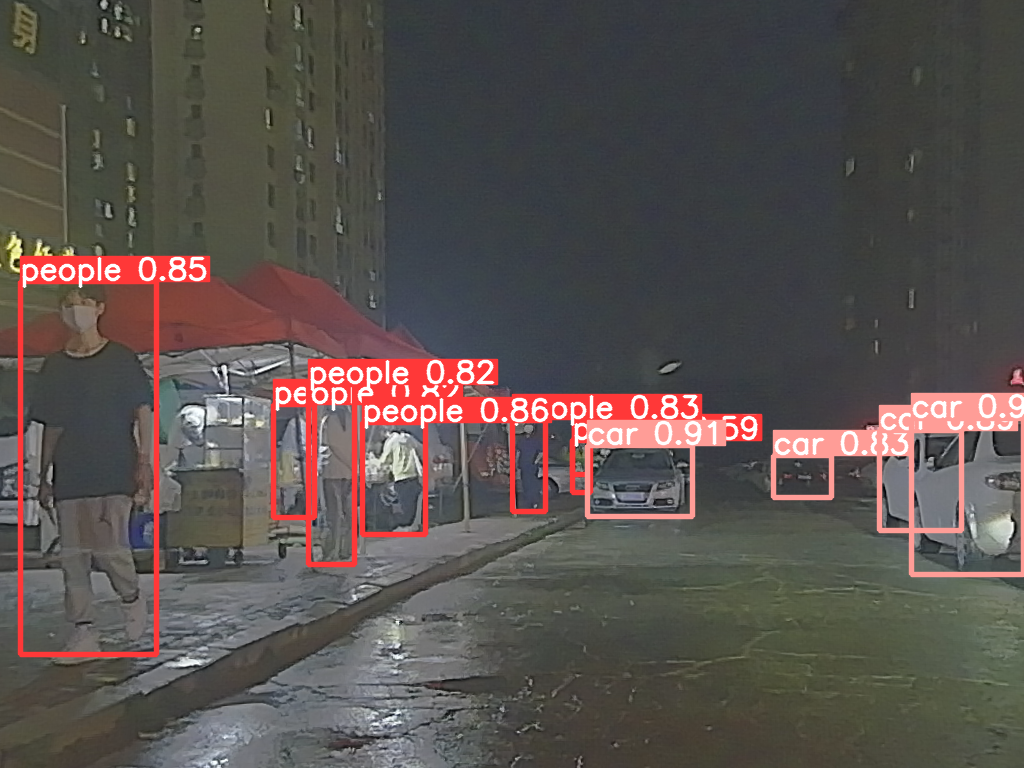}
        \caption{Visible}
    \end{subfigure}
    \begin{subfigure}{0.18\linewidth}
        \centering
        \includegraphics[width=\linewidth]{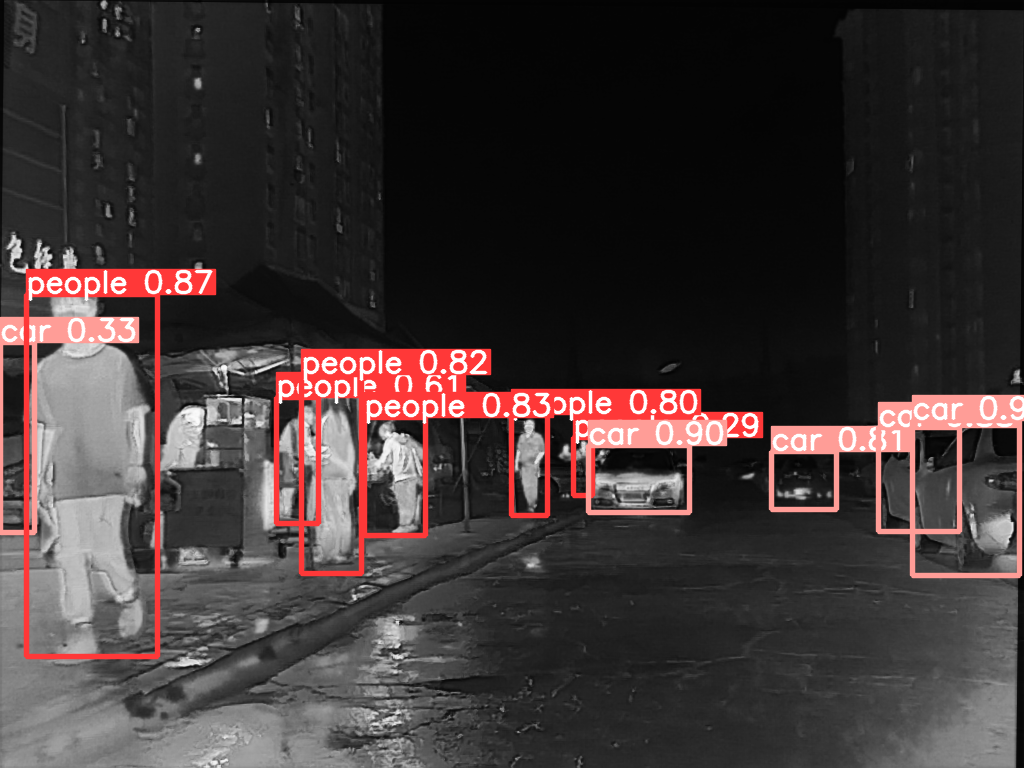}
        \caption{DIDFuse}
    \end{subfigure}
    \begin{subfigure}{0.18\linewidth}
        \centering
        \includegraphics[width=\linewidth]{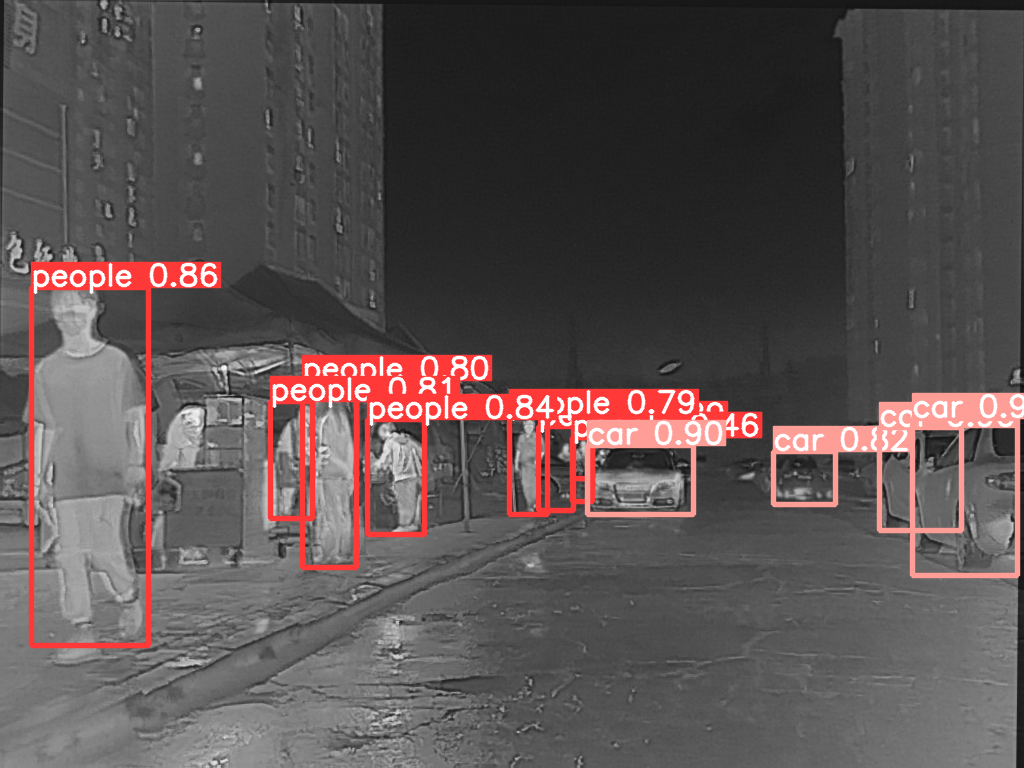}
        \caption{U2Fusion}
    \end{subfigure}
    \begin{subfigure}{0.18\linewidth}
        \centering
        \includegraphics[width=\linewidth]{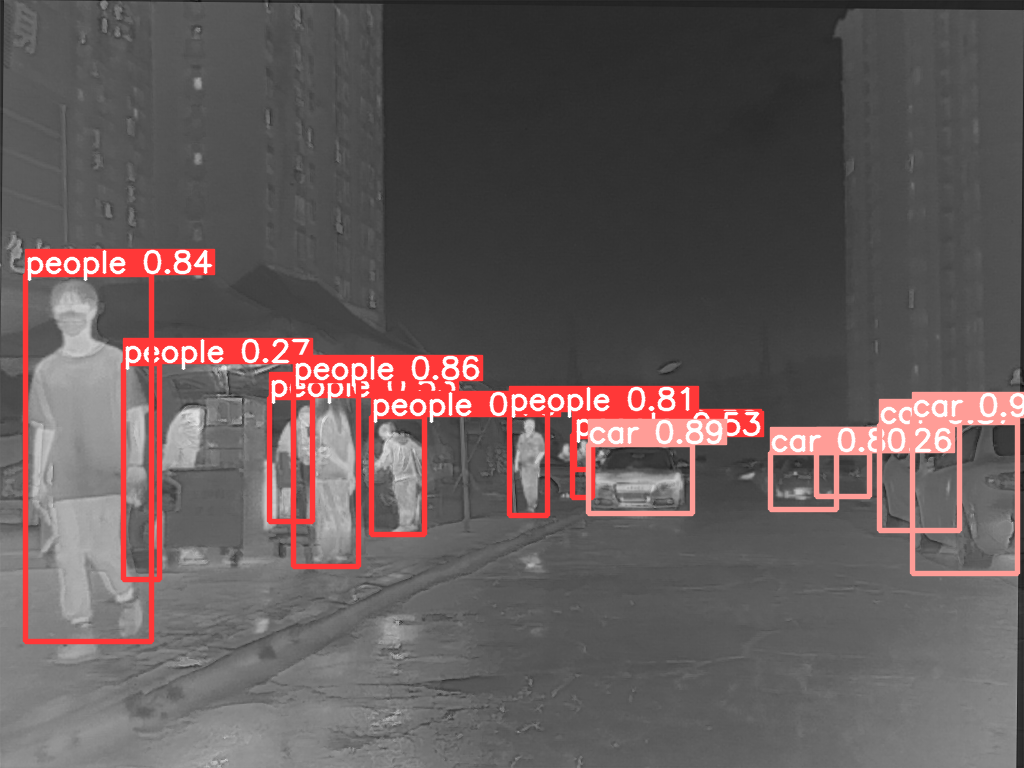}
        \caption{YDTR}
    \end{subfigure}

    \vspace{0.4cm}
    
    \begin{subfigure}{0.18\linewidth}
        \centering
        \includegraphics[width=\linewidth]{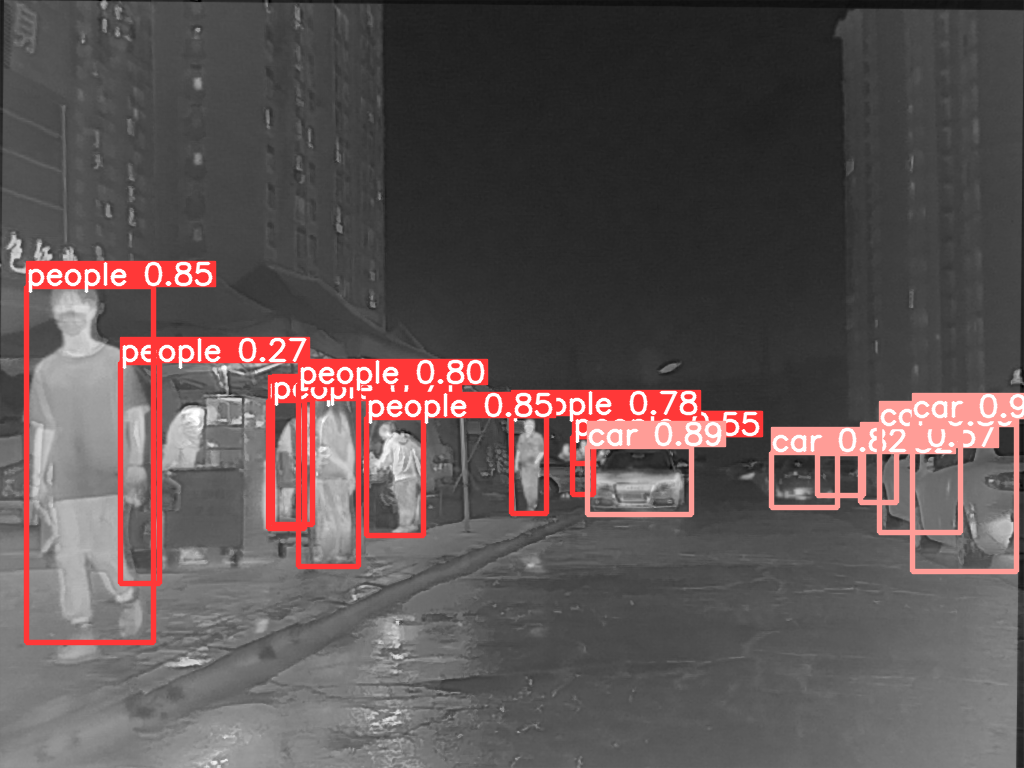}
        \caption{MFEIF}
    \end{subfigure}
    \begin{subfigure}{0.18\linewidth}
        \centering
        \includegraphics[width=\linewidth]{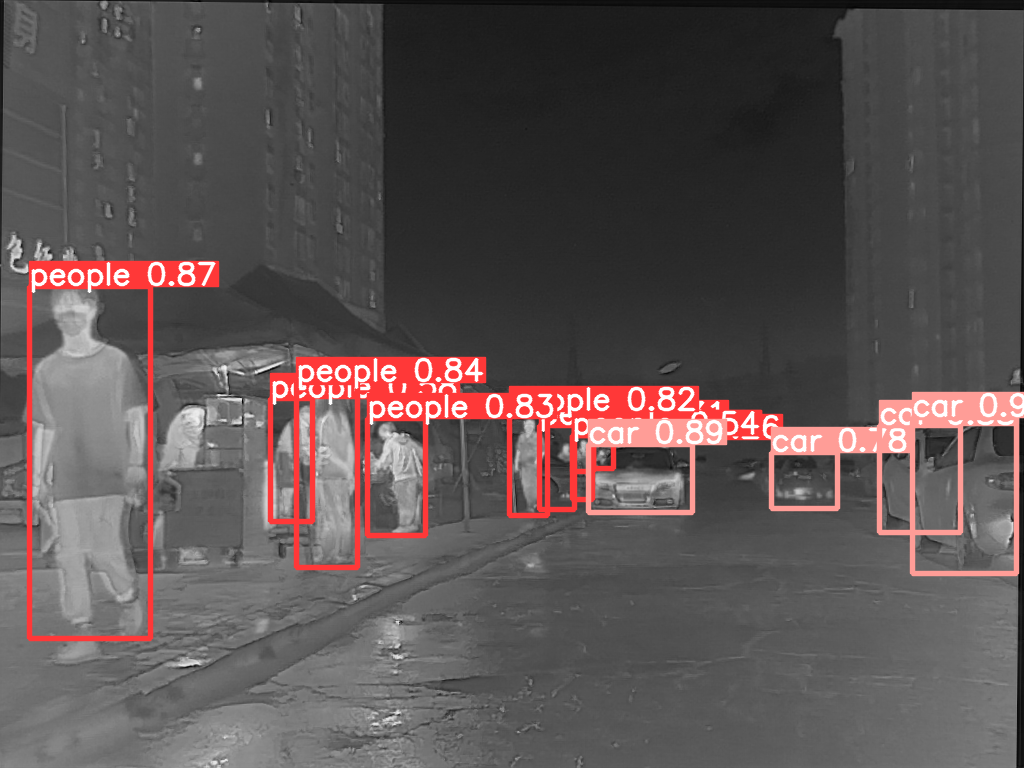}
        \caption{UMF-CMGR}
    \end{subfigure}
    \begin{subfigure}{0.18\linewidth}
        \centering
        \includegraphics[width=\linewidth]{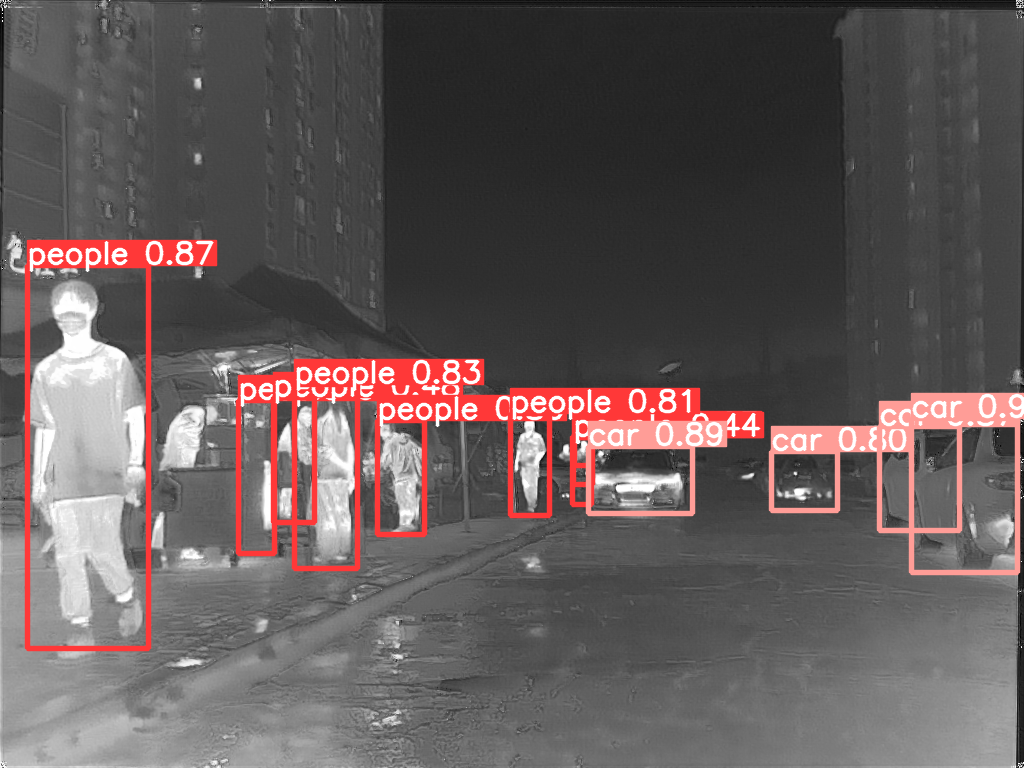}
        \caption{TarDal}
    \end{subfigure}
    \begin{subfigure}{0.18\linewidth}
        \centering
        \includegraphics[width=\linewidth]{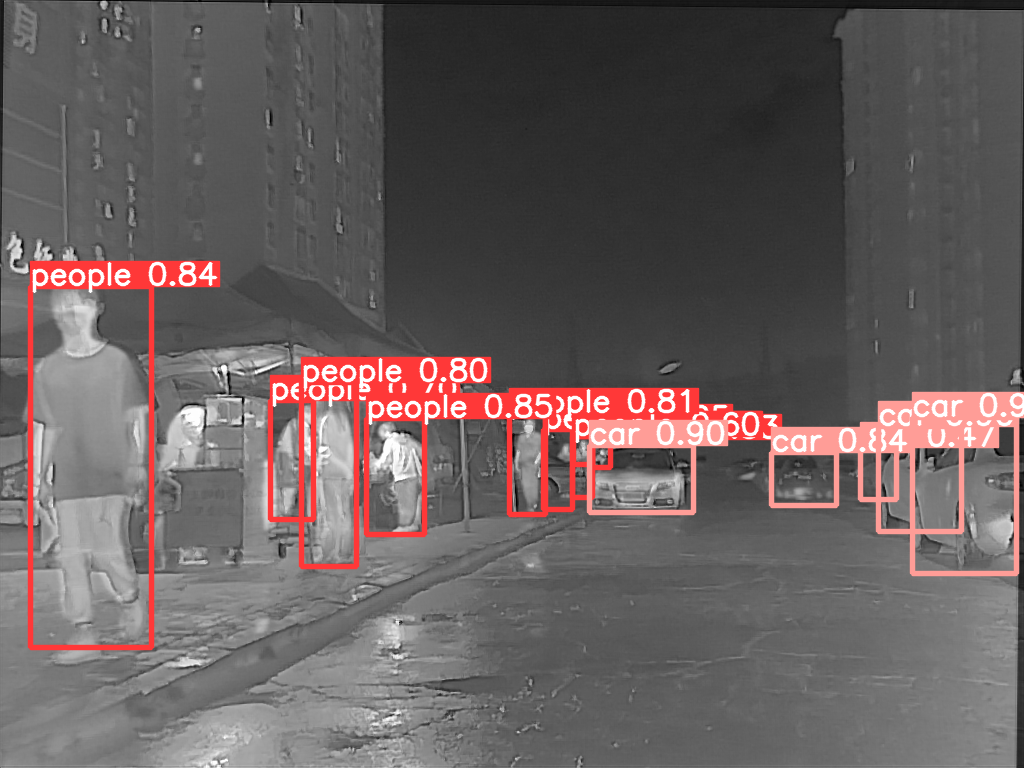}
        \caption{IRFS}
    \end{subfigure}
    \begin{subfigure}{0.18\linewidth}
        \centering
        \includegraphics[width=\linewidth]{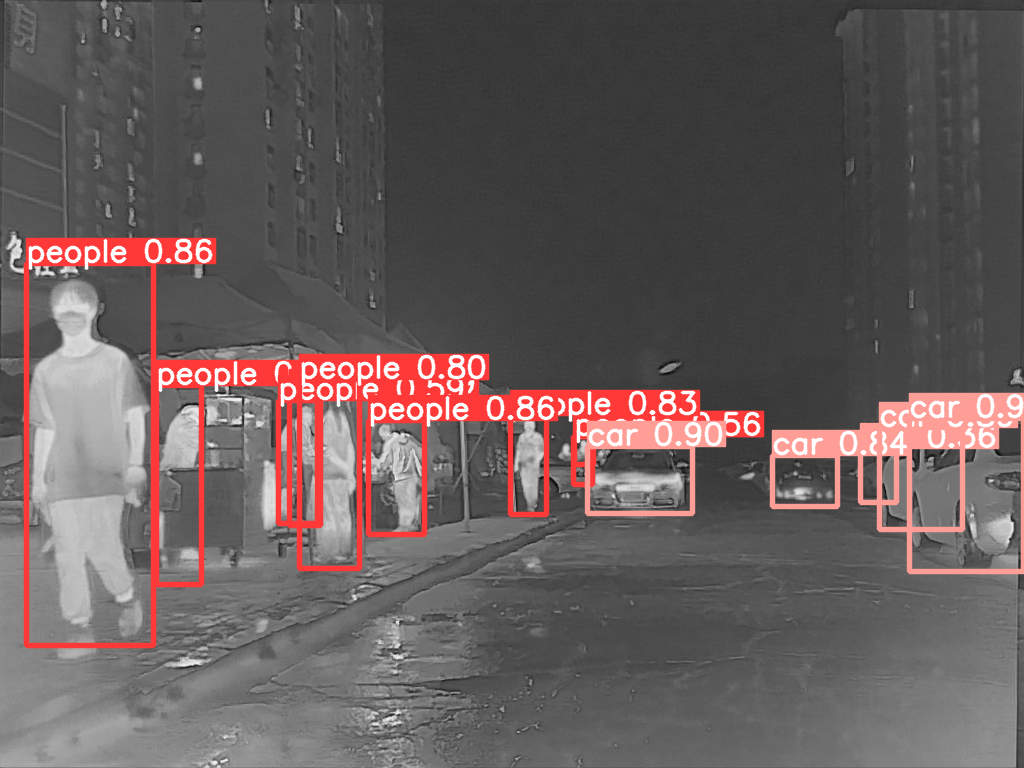}
        \caption{Ours}
    \end{subfigure}

    \vspace{0.4cm}
    
    \caption{A comparison of the detection performance after fusion on image "00548" from the M3FD dataset.}

    \vspace{0.4cm}
    
    \label{fig5}
\end{figure*}

Figure~\ref{fig5} intuitively demonstrates the detection results of fusion images generated by different methods. From the figure, it can be observed that for the person bending over behind the stall, the visible image as well as the images generated by methods such as DIDFuse, U2Fusion, UMF-CMGR, TarDal, and IRFS all exhibit instances of missed detections. Additionally, in the infrared images, the background is mistakenly identified as a person, while in the images generated by YDTR and MFEIF, the arms are incorrectly identified as people. Only in the image generated by our proposed SFDFusion, the target is accurately detected. This indicates that our fusion algorithm can effectively integrate valuable information from both the infrared and visible modalities, rather than solely enhancing the visual appearance of the images through other methods.

\begin{table}[t]
    \centering
    
    \caption{Results of ablation experiments on MSRS dataset.}

    \vspace{0.2cm}

    \label{tab3}
    \begin{tabular}{ccccccc}
        \toprule
        & EN & SD & SF & MI & VIF & Qabf \\
        \midrule
        w/o DMRM & 6.652 & 41.831 & 10.457 & \snd{3.094} & \snd{0.649} & 0.386 \\
        w/o FDFM & 6.641 & 41.847 & 10.058 & 2.948 & 0.589 & 0.390 \\
        w/o $\mathcal{L}_{fre}$ & \snd{6.657} & \snd{42.372} & \snd{10.567} & 3.000 & 0.630 & \snd{0.452}\\
        SFDFusion & \fst{6.670} & \fst{43.197} & \fst{11.070} & \fst{3.914} & \fst{1.011} & \fst{0.680} \\
        \bottomrule
    \end{tabular}
\end{table}

\subsection{Ablation Study}
We conduct ablation experiments on the MSRS dataset to validate the effectiveness of our proposed method and explore the influence of different components on fusion quality. Specifically, we individually remove the proposed DMRM, FDFM, and frequency domain fusion loss $(\mathcal{L}_{fre})$ to assess their impact. The results of the ablation experiments demonstrate the effectiveness of the module we proposed.

For the ablation experiment on DMRM, we remove the DMRM module and directly concatenate the fusion results of infrared and visible images in the frequency domain before inputting them into the fusion module. Regarding the ablation experiment on the $\mathcal{L}_{fre}$, we retained the complete network structure but excluded the $\mathcal{L}_{fre}$ from the total loss to investigate its impact on fusion quality. As for the ablation experiment on FDFM, we kept only the DMRM module and performed fusion solely using spatial domain information.

Table~\ref{tab3} shows the results of the ablation experiments, where "w/o" indicates the removal of a certain module. It can be observed that when any part is removed, the quality evaluation metrics of the fusion images generated by our method generally decrease. Among these metrics, MI, VIF, and Qabf are the most affected, indicating a degradation in the fusion results in terms of both the preservation of source image information and human visual perception effects. This suggests the effectiveness of the proposed modules and losses.

\begin{figure}[t]
    \centering
    \begin{subfigure}{0.4\linewidth}
        \centering
        \includegraphics[width=\linewidth]{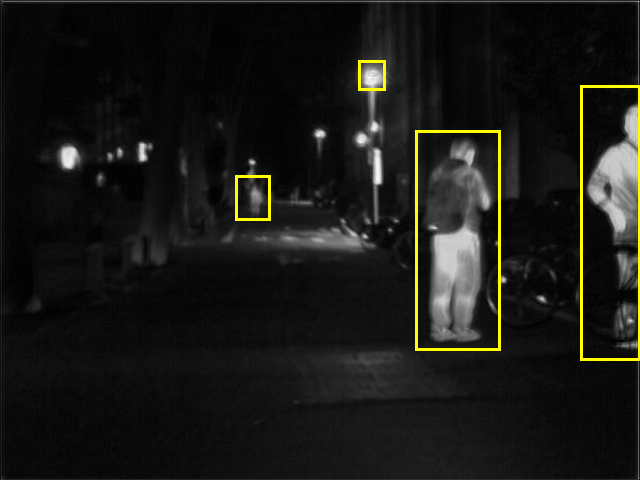}
        \caption{w/o DMRM}
    \end{subfigure}
    \hspace{0.2cm}
    \begin{subfigure}{0.4\linewidth}
        \centering
        \includegraphics[width=\linewidth]{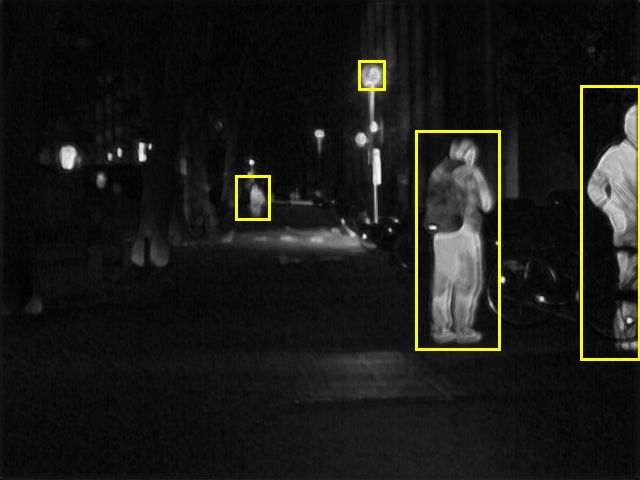}
        \caption{w/o FDFM}
    \end{subfigure}
    
    \vspace{0.3cm}
    
    \begin{subfigure}{0.4\linewidth}
        \centering
        \includegraphics[width=\linewidth]{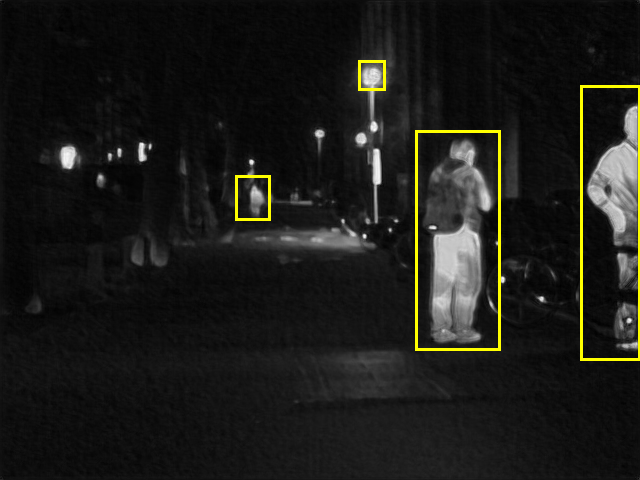}
        \caption{w/o $\mathcal{L}_{fre}$}
    \end{subfigure}
    \hspace{0.2cm}
    \begin{subfigure}{0.4\linewidth}
        \centering
        \includegraphics[width=\linewidth]{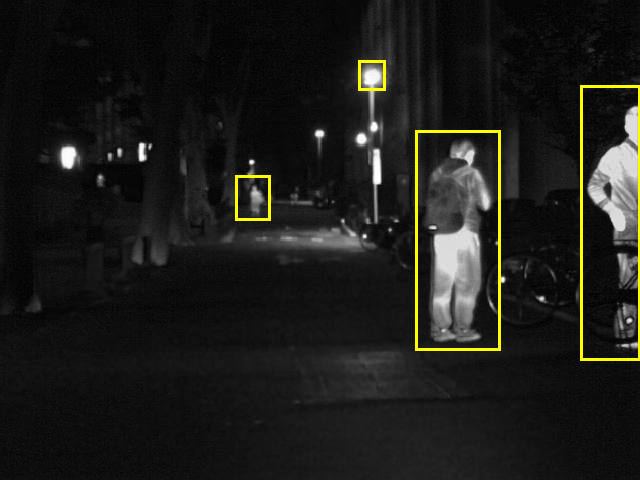}
        \caption{SFDFusion}
    \end{subfigure}

    \vspace{0.4cm}
    
    \caption{Visual comparison of ablation experiments.}
    \label{fig6}   

    \vspace{0.5cm}
    
\end{figure}

Figure~\ref{fig6} provides a visual representation of the different ablation experiments, with image "01012N" utilized. From the figure, it is evident that for the close-range targets within the yellow bounding box, our SFDFusion method maximally preserves the pixel intensity information from the infrared modality, while also retaining the texture information in the clothing of the individuals. Unlike other results where there is some degree of loss in both pixel intensity and texture. Regarding distant targets, SFDFusion effectively preserves the overall contour of the human body and enhances the brightness of lamp targets, achieving superior results.

\section{Conclusion}
In this paper, we introduced the SFDFusion network, a novel network for fusing infrared and visible images in the spatial domain and frequency domain. Distinguish with other fusion methods that incorporate frequency domain, our method captured effective information from the two modalities through a parallel structure. The proposed DMRM captures cross-modal feature information in the spatial domain to reconstruct and refine the original images. Additionally, another module called FDFM primarily fused frequency domain information. We also designed a frequency domain fusion loss to ensure consistency between the frequency domain and the spatial domain. Extensive experiments showed that our method has significant advantages in fusion metrics and visual effects. Moreover, our method demonstrated excellent efficiency in image fusion and better performance in downstream detection tasks.



\begin{ack}
This paper is supported in part by the Project of National Key Laboratory of Electromagnetic Energy (6142217040101), and in part by the Youth Talent Support Program of Beihang University (YWF-23-L-1244).
\end{ack}



\bibliography{mybibfile}

\end{document}